\documentclass{article}

\usepackage{PRIMEarxiv}

\usepackage[utf8]{inputenc} 
\usepackage[T1]{fontenc}    
\usepackage{hyperref}       
\usepackage{url}            
\usepackage{booktabs}       
\usepackage{amsfonts}       
\usepackage{nicefrac}       
\usepackage{microtype}      
\usepackage{lipsum}
\usepackage{fancyhdr}       
\usepackage{graphicx}       

\usepackage{multirow}
\usepackage{amsmath}
\usepackage{xcolor}
\usepackage{arydshln} 
\usepackage{blindtext}
\usepackage[switch]{lineno}
\usepackage{subcaption}
\usepackage{varwidth}
\newsavebox\tmpbox
\colorlet{color1}{green!10!orange!90!}
\usepackage{footnote}

\title{Overcoming Difficulty in Obtaining Dark-skinned Subjects for Remote-PPG by Synthetic Augmentation}

\newcommand*\samethanks[1][\value{footnote}]{\footnotemark[#1]}
\author{
  Yunhao Ba$^{1}$\thanks{Equal contribution. Order randomly determined.}, Zhen Wang$^{1}$\samethanks{}, Kerim Doruk Karinca$^{2}$, Oyku Deniz Bozkurt$^{1}$, Achuta Kadambi$^{3}$\thanks{Corresponding author.}\\
  University of California, Los Angeles \\
  $^{1}$\texttt{\{yhba, zhenwang, denizbozkurt\}@ucla.edu}, $^{2}$\texttt{dorukkarinca@cs.ucla.edu}, $^{3}$\texttt{achuta@ee.ucla.edu}\\
}

\begin{document}
\maketitle

\begin{abstract}
  Camera-based remote photoplethysmography (rPPG) provides a non-contact way to measure physiological signals (e.g., heart rate) using facial videos. Recent deep learning architectures have improved the accuracy of such physiological measurement significantly, yet they are restricted by the diversity of the annotated videos. The existing datasets MMSE-HR, AFRL, and UBFC-RPPG contain roughly 10\%, 0\%, and 5\% of dark-skinned subjects respectively. The unbalanced training sets result in a poor generalization capability to unseen subjects and lead to unwanted bias toward different demographic groups. In Western academia, it is regrettably difficult in a university setting to collect data on these dark-skinned subjects. Here we show a first attempt to overcome the lack of dark-skinned subjects by synthetic augmentation. A joint optimization framework is utilized to translate real videos from light-skinned subjects to dark skin tones while retaining their pulsatile signals. In the experiment, our method exhibits around 31\% reduction in mean absolute error for the dark-skinned group and 46\% improvement on bias mitigation for all the groups, as compared with the previous work trained with just real samples.
\end{abstract}


Heart rate along with other cardiovascular parameters is identified as an independent risk factor for cardiovascular disease~\cite{fox2007resting, perret2009heart}. During the pandemic, telehealth consults have increased more than 50-fold for certain groups (e.g., those with chronic diseases)~\cite{weiner2021person} due to the concerns that the congregation of people may increase the risk of contraction. Although contact sensors (electrocardiograms, oximeters) provide gold-standards for measuring heart function~\cite{liu2021metaphys}, these contact-devices are not widely available. Therefore, a non-contact way of detecting vital signs is crucial for the telehealth settings. Non-contact health sensing can also benefit applications in the clinical settings, such as neonatal ICU sensing, as the contact sensors may cause infection for these vulnerable groups. Camera-based remote photoplethysmography (rPPG) method~\cite{verkruysse2008remote} provides a solution to the above scenarios as web cameras are more ubiquitously available, contactless, and low-cost. It uses subtle skin color variations on the face to obtain the physiological signals. When the light hits the face, the amount of light reflected or absorbed will be determined by the physiological processes, and the color change corresponding to the Blood Volume Pulse (BVP) is synchronized with the heart rate (HR), which provides the feasibility to extract HR from facial videos.

\begin{figure}[t]
    \centering
    \includegraphics[width=0.6\columnwidth]{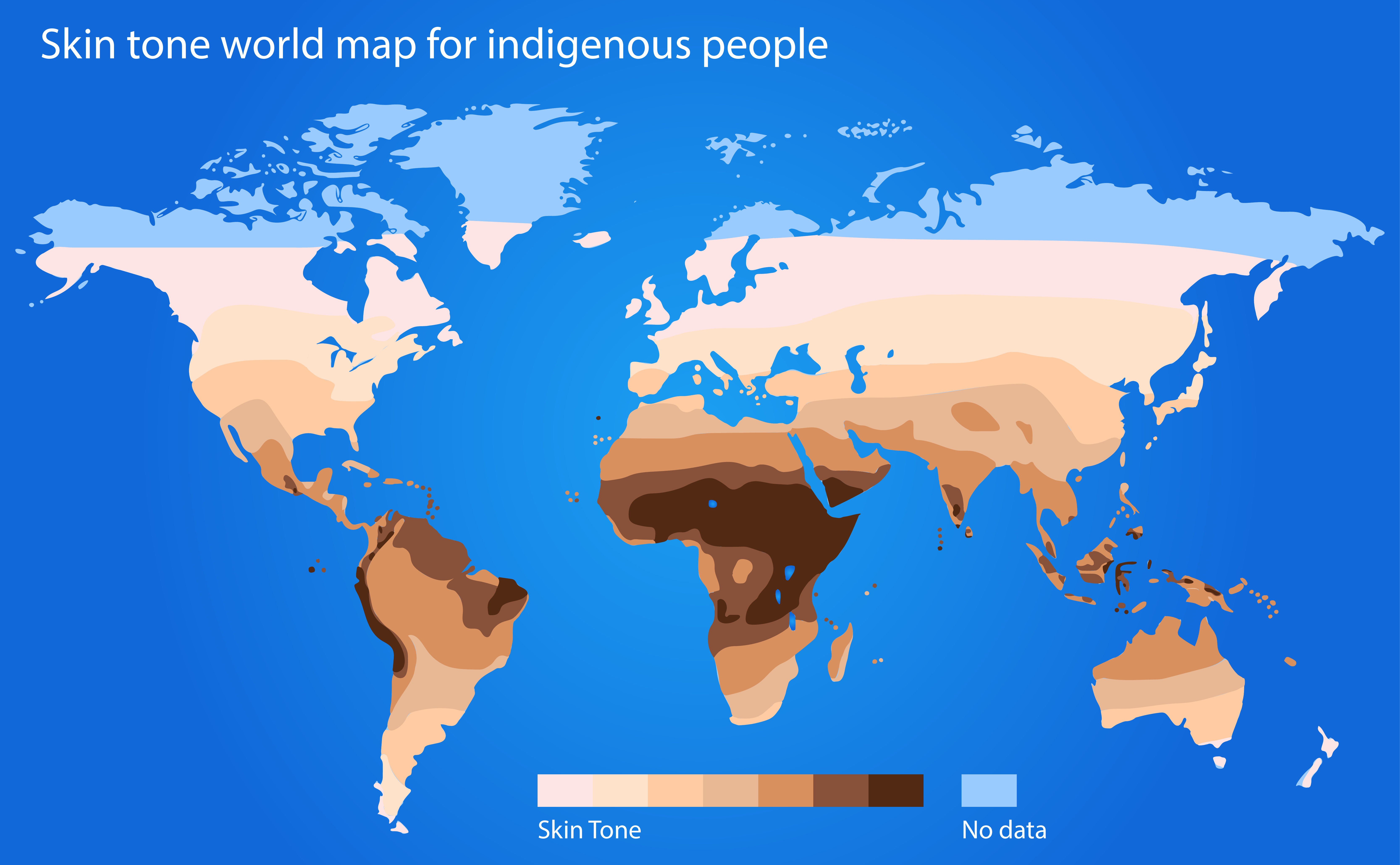}
    \caption{\textbf{Skin color reflectance map for indigenous people.} A diverse rPPG dataset may not be accessible for some countries/regions due to the skin color distribution. Synthetic dark-skinned subjects are critical for the worldwide deployment of rPPG. Skin color data from Chaplin, G.~\cite{chaplin2004geographic}.}
    \label{fig:map}
\end{figure}

\begin{figure*}[t]
    \centering
    \includegraphics[width=\textwidth]{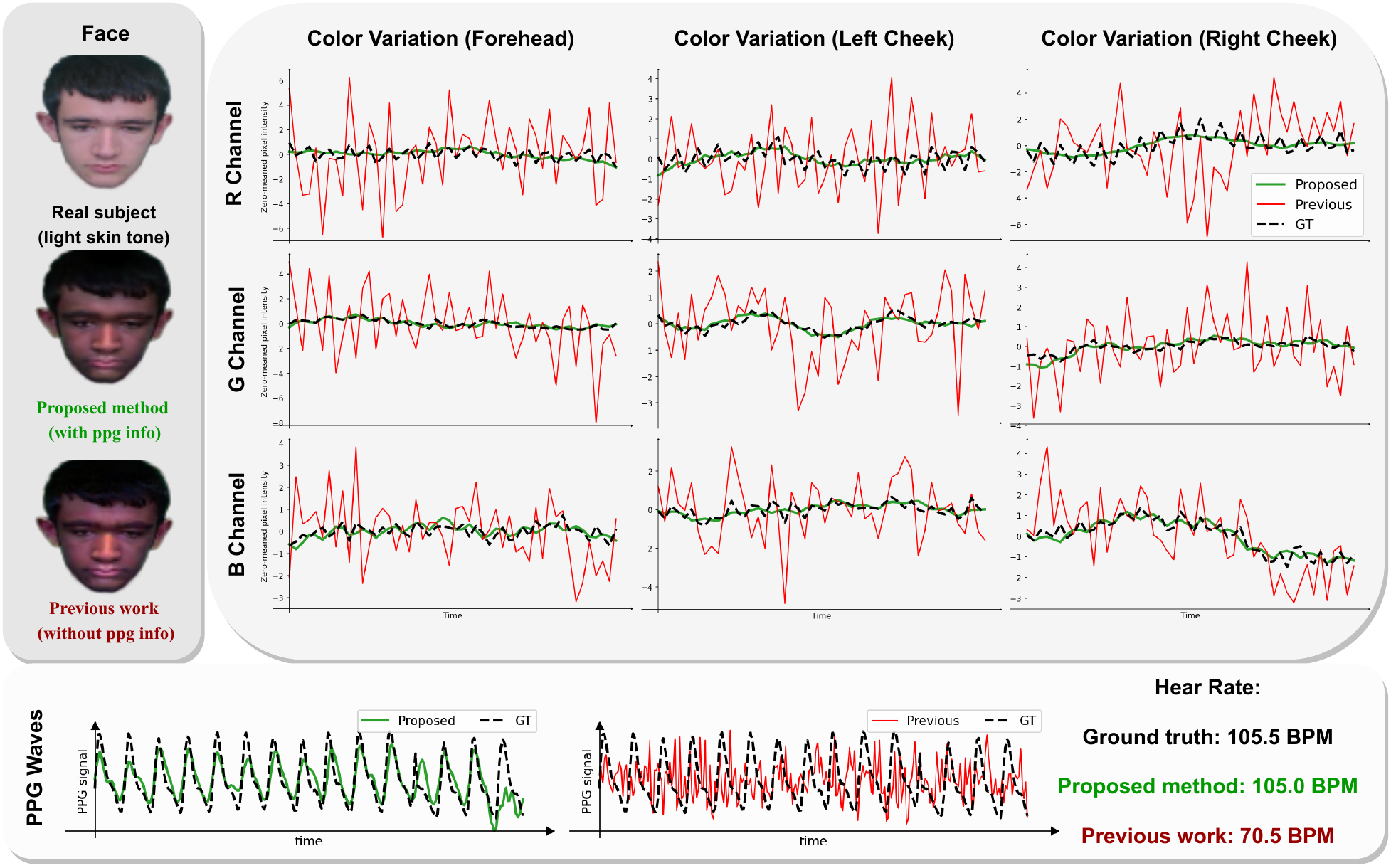}
    \caption{\textbf{The proposed method successfully incorporates pulsatile signals into the generated videos, while the existing work~\cite{yucer2020exploring} only focuses on the visual appearance.} For different facial regions, frames generated by the proposed method exhibit similar pixel intensity variations as compared with frames from real videos, while the prior work shows unrealistic RGB variations. As a result, pulsatile signals can be well preserved in our method as opposed to the vanilla skin tone translation.}
    \label{fig:teaser}
\end{figure*}

Algorithms for non-contact rPPG can be roughly classified into three categories: Signal decomposition~\cite{lewandowska2011measuring, poh2010advancements, poh2010non, tulyakov2016self, wang2015novel}, model-based methods~\cite{de2014improved, de2013robust, wang2016algorithmic}, and deep learning methods~\cite{mcduff2018deep, yu2019remote, reiss2019deep, niu2019rhythmnet}. Signal decomposition techniques based on Blind Source Separation (BSS) decompose/demix the face videos into different sources utilizing PCA~\cite{lewandowska2011measuring} or ICA~\cite{poh2010advancements, poh2010non}. For model-based methods, Pulse Blood Vector~\cite{de2014improved} utilizes the characteristic blood volume signature to weight different color channels. CHROM~\cite{de2013robust} first eliminates the specular components and applies color space transforms to linearly combine the chrominance signals. POS~\cite{wang2016algorithmic} modifies this by first projecting the temporally-normalized skin tone onto the plane which is orthogonal to the intensity variation direction and then linearly combine the projected signals. These model-based methods usually use spatially averaged intensity values of skin pixels for pulse extraction, which may achieve sub-optimal results as each pixel can contribute differently to the underlying pulse signals.

While data-driven neural networks have exhibited remarkable estimation accuracy for non-contact camera-based sensing~\cite{mcduff2018deep, yu2019remote, reiss2019deep, niu2019rhythmnet}, there exist several practical constraints towards collecting large-scale data from patients for these deep learning models: (1) demographic biases in society that translate to data (e.g., innovation happening in some countries/regions may not have access to a diverse dataset as illustrated in Figure~\ref{fig:map}); (2) the requirement of medical-grade sensors and necessity of intrusive/semi-intrusive traditional methods for data collection; (3) patient privacy concerns (e.g., OBF dataset~\cite{li2018obf} is not publicly available due to the licence issue). 

Recent study shows that computer vision algorithms have been disadvantaging the underrepresented groups in some applications, such as face recognition~\cite{buolamwini2018gender}. Non-contact rPPG estimation is not an exception given the unbalanced and relatively small datasets in the field~\cite{nowara2020meta}. There are very rare subjects with dark skin tones in the existing benchmark datasets. More specifically, MMSE-HR~\cite{zhang2016multimodal}, AFRL~\cite{estepp2014recovering}, and UBFC-RPPG~\cite{bobbia2019unsupervised} only contain roughly 10\%, 0\%, and 5\% dark-skinned subjects respectively. With the training sets heavily biased towards subjects of light skin tones, the state-of-the-art data-driven rPPG models usually fail to generalize their performance to the underrepresented groups~\cite{nowara2020meta}. This prohibits the clinical deployment of these algorithms, since it is critical for rPPG algorithms to have consistent performance across different demographic groups in the clinical settings. 

Realizing the difficulty of recruiting patients to collect large-scale rPPG datasets in the university setting, synthetic augmentation of facial videos has become an active research topic recently. McDuff et al.~\cite{mcduff2020advancing} use synthetic avatars with ray tracing to reflect the blood volume changes under various configurations. However, as the authors point out, that infrastructure is labor-intensive and requires a significant amount of rendering time for each frame (approximately 20 seconds per frame), which impedes their scalability. Pulse signals can also be incorporated to make the synthetic avatars more lifelike, yet it is difficult for avatar-based methods to generate a balanced dataset due to the lack of dark-skinned avatars~\cite{mcduff2021warm}. Tsou et al.~\cite{tsou2020multi} augment source rPPG videos with other specified pulse signals. However, their framework is restricted to the face appearance in the original source videos and fails to produce novel videos with dark skin tones.

In contrast to these prior arts, we do a first attempt to directly augment the existing rPPG dataset by translating videos of light-skinned subjects to dark skin tones. This is difficult because the color variations due to blood volume changes are subtle, and the generation network has to be carefully designed to reflect these subtle changes while conducting skin tone translation without accessing real rPPG videos of dark-skinned subjects. However, this technique is rewarding, since it is capable of producing both photo-realistic and physiologically accurate synthetic videos in a fast manner (approximately 0.005 seconds per frame in average for our model) and can assist the development of algorithms and techniques for remote diagnostics and healthcare. In the experiment, our proposed method can reduce around 31\% HR estimation error for the dark-skinned group and show 46\% improvement on bias mitigation for all the groups, as compared with the existing architecture trained with just real samples.

Yucer et al.~\cite{yucer2020exploring} introduce a race translation model across various racial domains with a CycleGAN-based architecture. However, their work is not designed to incorporate pulsatile signals. As illustrated in Figure~\ref{fig:teaser}, this vanilla skin tone translation network~\cite{yucer2020exploring} merely focuses on the visual appearance, and the pulsatile signals are not preserved. To address this issue, we propose a learning framework that can augment realistic rPPG videos with dark skin tones that are of high fidelity. The framework consists of two interconnected components: (1) a generator to translate light skin tones to dark skin tones and (2) an rPPG estimator named PhysResNet (PRN) to encourage pulsatile signals within the generated videos. The generator is trained to learn both the visual appearance and the subtle color variations with respect to the underlying blood volume variations, and the rPPG network can simultaneously benefit from the generator to generalize its performance in diverse groups. We also demonstrate that our generated synthetic videos can be directly utilized to improve the performance of the state-of-the-art data-driven rPPG method with reduced bias across different skin color groups.

\begin{figure*}[t]
    \centering
    \includegraphics[width=\textwidth]{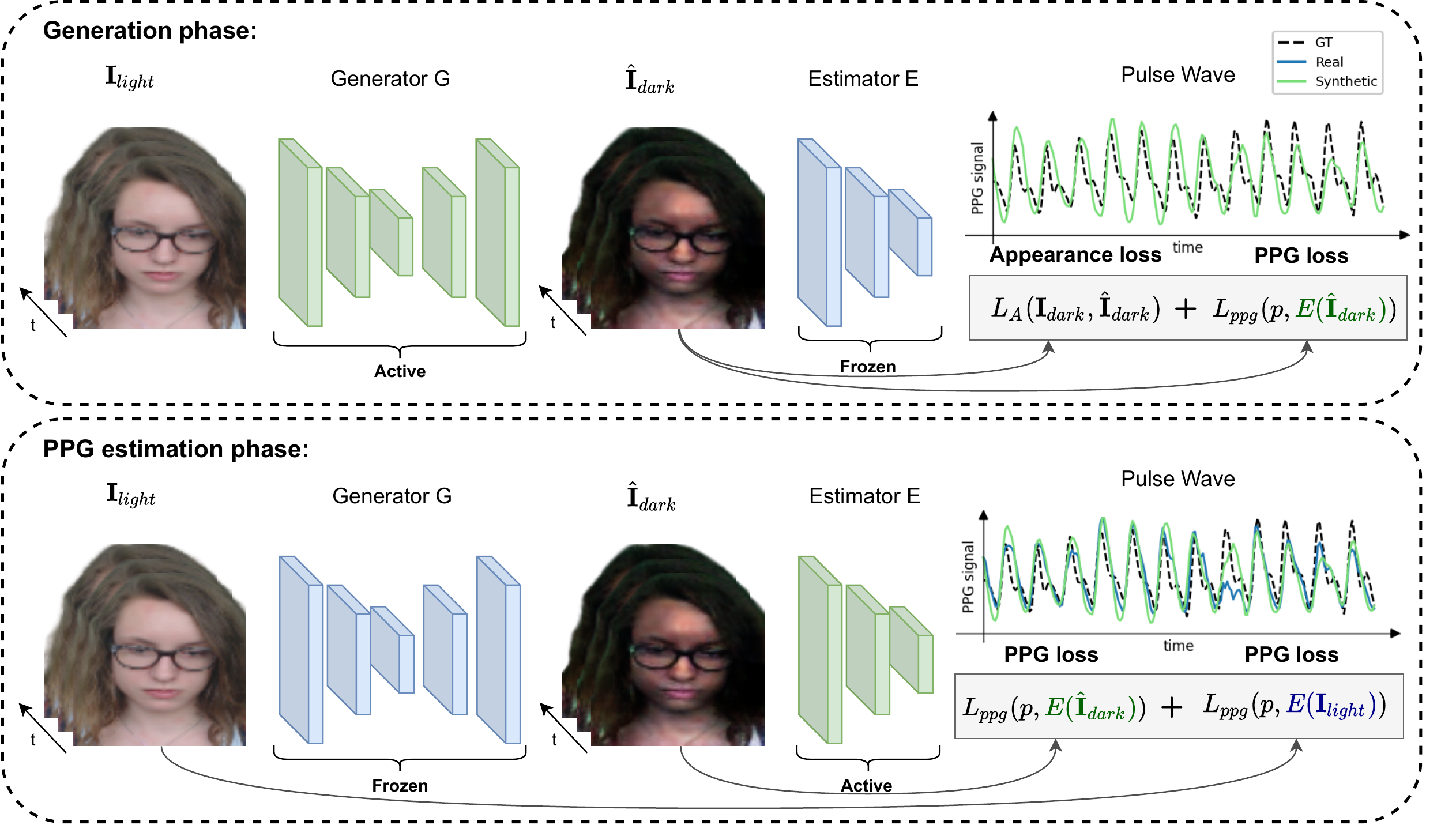}
    \caption{\textbf{Illustration of the proposed joint optimization framework.} Our framework is capable of translating light-skinned facial videos to dark skin tones while maintaining the original pulsatile signals. With a two-phase weight updating scheme, the rPPG estimation network can benefit from the synthetic dark-skinned videos and gradually learn to conduct inference on dark-skinned subjects without accessing real facial videos with dark skin tones.}
    \label{fig:pipeline}
\end{figure*}

\section*{Results}

\subsection*{Bio-realistic skin tone translation} \label{sec:method}

In order to translate real subjects with light skin tones to synthetic subjects with dark skin tones, we utilize two interconnected networks: a video generator $G$ and an rPPG estimator $E$, as illustrated in Figure~\ref{fig:pipeline}. We next describe the proposed 3D convolutional video generator, the rPPG estimation network, and our joint optimization scheme.

\subsubsection*{3D convolutional video generator} \label{sec:generator}

The goal of our video generator $G$ is to translate frame sequences of real light-skinned subjects to synthetic dark-skinned subjects. We propose a novel 3D convolutional neural network to accomplish this goal. The model consists of an encoder (several convolutional layers), a transformer (6 ResNet Blocks), and finally a decoder (several convolutional layers). Please refer to the supplementary material for a detailed description of the network architecture.

The generator takes 256 consecutive frames $\textbf{I}_{light}$ at size $80 \times 80$ as the input and generates the corresponding translated frames in the same dimension. Since the paired ground-truth translated frames do not exist, we use a race transfer model~\cite{yucer2020exploring} pretrained on VGGFace2~\cite{cao2018vggface2} to generate the pseudo target frames $\textbf{I}_{dark}$. More specifically, the generator \textit{Caucasian-to-African} in~\cite{yucer2020exploring} is utilized to translate videos of light-skinned subjects in the existing rPPG dataset to dark skin tones.

The generator is first supervised by the L1 distance between the pseudo target frames $\textbf{I}_{dark}$ and the generated frames $\hat{\textbf{I}}_{dark} = G(\textbf{I}_{light})$ to learn the visual appearance of the synthetic dark-skinned subjects. At this stage, the output frames $\hat{\textbf{I}}_{dark}$ do not contain pulsatile signal, since the target frames $\textbf{I}_{dark}$ from~\cite{yucer2020exploring} are generated in a frame-by-frame manner without temporal pulse correspondence along the time dimension. In the joint optimization part, we describe how to further incorporate the pulsatile signals presented in the original videos $\textbf{I}_{light}$ into the generated frames.  

\subsubsection*{PRN: rPPG estimator with residual connections} \label{sec:rppg_net}

The rPPG estimator is designed to model the BVP temporal information from a sequence of facial frames. Similarly, it takes 256 consecutive frames at size $80 \times 80$ as the input, and its output is the corresponding BVP value for each input frame. We build our novel rPPG estimator based on 3D convolution operations. It consists of three consecutive 3D convolutional blocks with residual connections, and an average pooling is performed after each block for the downsampling purpose. A detailed description of each block can be found in the supplementary material. 

To supervise the network, we use a negative Pearson correlation loss between the estimated pulse signals $\hat{p} \in \mathbb{R}^T$ and the ground-truth pulse signals $p \in \mathbb{R}^T$:
\begin{equation}
\begin{aligned}
L&_{ppg}(p, \hat{p}) = 1 - \\
&\frac{T \sum_{i=1}^{T} p_{i} \hat{p}_{i}-\sum_{i=1}^{T} p_{i} \sum_{i=1}^{T} \hat{p}_{i}}{\sqrt{\left(T \sum_{i=1}^{T} p_{i}^{2}-\left(\sum_{i=1}^{T} p_{i}\right)^{2}\right)\left(T \sum_{i=1}^{T}\left(\hat{p}_{i}\right)^{2}-\left(\sum_{i=1}^{T} \hat{p}_{i}\right)^{2}\right)}}.
\end{aligned}
\end{equation}
This negative Pearson correlation loss has shown to be more effective as compared with the point-wise mean squared error (MSE) loss in the previous work~\cite{yu2019remote}. We first train PRN with only real subjects, and this simple yet efficient architecture can already achieve state-of-the-art performance on the existing rPPG datasets. In next part, we detail how to further incorporate the synthetic subjects into the training process.

\subsubsection*{Joint optimization} \label{sec:joint_learning}

The generator trained with L1 loss in the previous part fails to produce synthetic dark-skinned subjects with desired pulsatile information, and the rPPG estimator trained with only real light-skinned subjects exhibits poor generalization capability on unseen data or data that rarely appears in the training set (i.e., the underrepresented group with dark skin tones). To make use of these two models, we design a joint optimization mechanism to incorporate pulsatile signals into the synthetic videos and improve the generalizability of the rPPG estimator simultaneously. 

We use a two-phase weight updating scheme to train the video generator and the rPPG estimator simultaneously. These two phases are alternated within each mini-batch as illustrated in Figure~\ref{fig:pipeline}. In the generation phase, we freeze the weight of the rPPG estimator $E$, and the generator $G$ is supervised by the following loss function to maintain both the visual appearance and the pulsatile information:
\begin{gather}
    L_G(\textbf{I}_{light}, p) =  L_{ppg}(p, E(\hat{\textbf{I}}_{dark})) + \lambda * L_{A}(\textbf{I}_{dark}, \hat{\textbf{I}}_{dark}), \\
    L_{A}(\textbf{I}_{dark}, \hat{\textbf{I}}_{dark}) = \frac{1}{\sum_i z_i} \sum_i z_i |\textbf{I}_{dark_i} - \hat{\textbf{I}}_{dark_i}|,  \\\quad z_i = 
        \begin{cases}
            0 & \text{if $|\textbf{I}_{dark_i} - \hat{\textbf{I}}_{dark_i}| < \epsilon$}\\
            1 & \text{otherwise}\\
        \end{cases},
\end{gather}
where $\hat{\textbf{I}}_{dark} = G(\textbf{I}_{light})$ is the generated frame sequence from synthetic dark-skinned subjects, $\lambda$ is the balance factor, $L_A(\cdot)$ is the visual appearance loss designed based on a threshold L1 loss, and $\epsilon$ is the selected threshold. The weighting factor $\lambda$ is chosen to be 1.0. Directly enforcing a L1 loss between $\textbf{I}_{dark}$ and $\hat{\textbf{I}}_{dark}$ causes the generator to struggle between the visual appearance and the pulse information, since the pseudo ground-truth $\textbf{I}_{dark_i}$ from~\cite{yucer2020exploring} do not contain the desired BVP variations. Therefore, we relax the appearance loss $L_A(\cdot)$ by a threshold $\epsilon$. The relaxation is based on the observation that the color changes due to BVP variations are subtle in the RGB domain. In our implementation, we select $\epsilon = 0.1$ based on an empirical analysis of the color variations in real videos.

In the rPPG estimation phase, we freeze the weight of the generator $G$ and train the rPPG estimator $E$ with both real and synthetically augmented frame sequences:
\begin{equation} 
    L_E(\textbf{I}_{light}, \hat{\textbf{I}}_{dark}), p) = L_{ppg}(p, E(\hat{\textbf{I}}_{dark})) + L_{ppg}(p, E(\textbf{I}_{light})).
\end{equation}
Both real and synthetic subjects are utilized to supervise the rPPG network $E$ while updating its weights. This arrangement allows $E$ to gradually adapt to the synthetic dark-skinned subjects without losing estimation accuracy on real subjects. With this two-phase updating rule, both the generator and the rPPG estimator benefit from each other in an alternate manner. At convergence, the generator $G$ can successfully translate frame sequences from real light-skinned subjects to dark skin tones while maintaining the original BVP variations, and the estimator $E$ can generalize its performance to dark skin tones without using actual real videos from dark-skinned subjects. 

\subsection*{Generating synthetic subjects with dark skin tones}

We demonstrate the superiority of our proposed method with empirical results on UBFC-RPPG~\cite{bobbia2019unsupervised} and VITAL~\cite{chari2020diverse} for HR estimation using various metrics: mean absolute error (MAE), root mean square error (RMSE), Pearson’s correlation coefficient (PCC), and signal-to-noise ratio (SNR). Please refer the method section for a detailed description of the datasets and the metrics. The synthetic videos generated by our model can also further improve the performance of the existing data-driven PPG estimation model with reduced bias across different skin tones.  

\begin{figure*}[t]
    \centering
    \includegraphics[width=\textwidth]{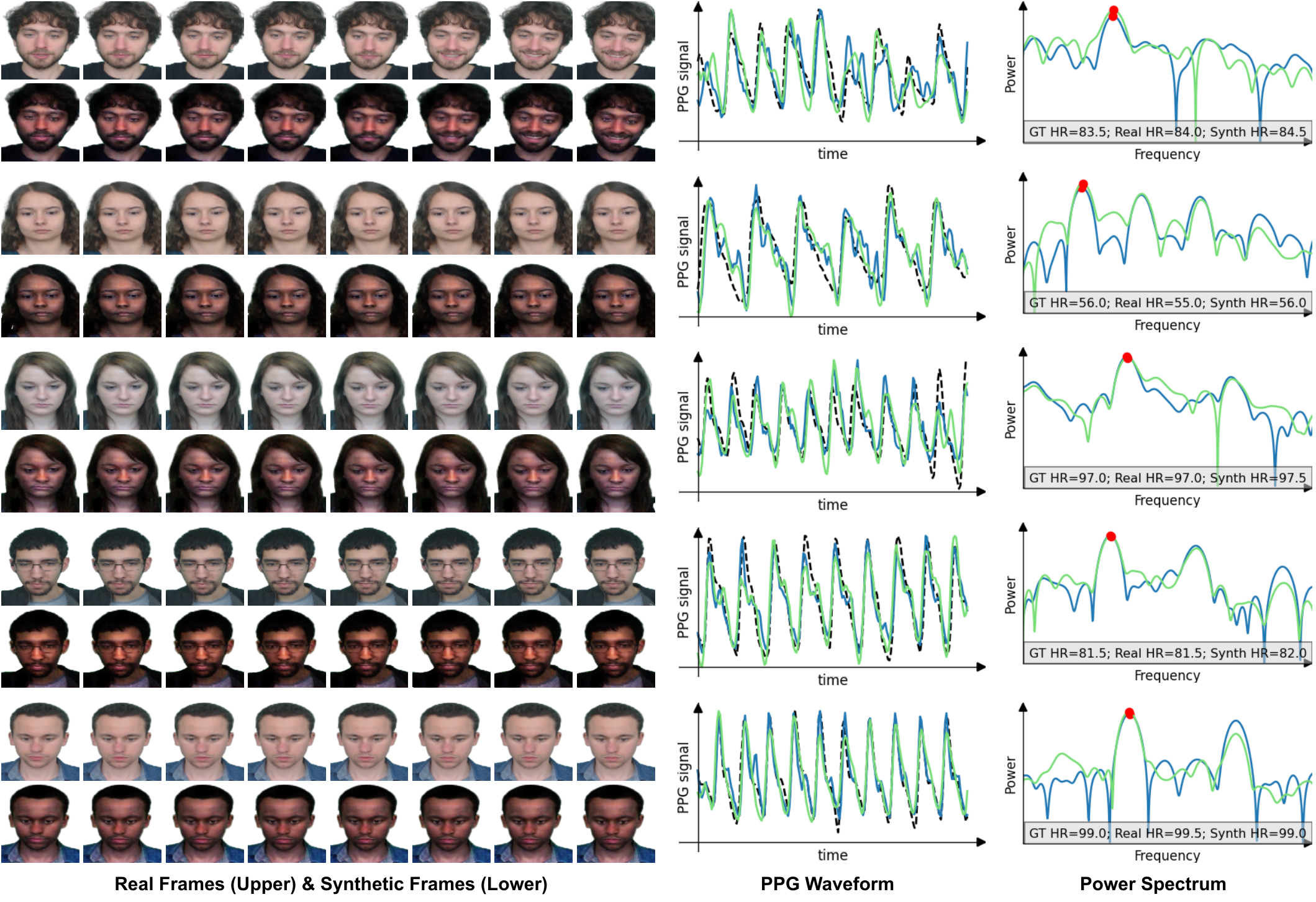}
    \caption{\textbf{Illustration of real frames and the corresponding synthetic frames in the UBFC-RPPG dataset.} Our proposed framework has successfully incorporated pulsatile signals when translating the skin color. The estimated pulse waves from PRN exhibit high correlation to the ground-truth waves, and the heart rates are preserved in the frequency domain.}
    \label{fig:darker_skin_results}
\end{figure*}

UBFC-RPPG dataset is randomly split into a training set (32 subjects) and a validation set (10 subjects). The training set is used to jointly optimize the generator $G$ and the rPPG estimator $E$. Models with minimum validation loss are selected for a cross-dataset evaluation on the VITAL videos. Some generated frames in the UBFC-RPPG validation set are illustrated in Figure~\ref{fig:darker_skin_results}. Our generator $G$ can successfully produce photo-realistic videos that reflect the associated underlying blood volume changes. Estimated pulse waves from the real videos and the synthetic videos are both closely aligned with the ground truth. In the frequency domain, power spectrum of the PPG waves is also preserved with a clear peak near the gold-standard HR value. 

\subsection*{Performance on UBFC-RPPG} \label{sec:ubfc}
\begin{table}[t]
\small
  \caption{\textbf{Performance of HR estimation on UBFC-RPPG.} Boldface font represents the preferred results.}
    \label{tab:ubfc}
  \begin{center}
  \begin{tabular}{lccccccc} 
    \toprule
  
    Method & MAE & RMSE & PCC & SNR  \\
    \midrule
    PRN augmented & \textbf{0.68} & \textbf{1.31} & 0.86 & 5.76 \\
  PRN w/ Real & 0.75 & 1.64 & 0.83 & \textbf{7.91} \\
  PRN w/ Synth & 4.32 & 6.56 & 0.54 & -1.93 \\
    3D-CNN~\cite{tsou2020multi} w/ Real$\&$Synth &  0.89 & 1.66 & \textbf{0.88} & 7.74\\
    3D-CNN~\cite{tsou2020multi} w/ Real  &1.09 & 1.91 & 0.84 & 7.80 \\
    3D-CNN~\cite{tsou2020multi} w/ Synth & 0.95 & 1.80 & 0.82 & 3.48 \\
   \hdashline
   POS~\cite{wang2016algorithmic} &3.69 & 5.31 & 0.75 & 3.07 \\
   CHROM~\cite{de2013robust} &1.84 & 3.40 &0.77 & 4.84 \\
   ICA~\cite{poh2010advancements} &8.28 &9.82 & 0.55 & 1.45\\
    \bottomrule
  \end{tabular}
 \end{center}
\end{table}

Performance metrics of different models in the UBFC-RPPG validation set are listed in Table~\ref{tab:ubfc}. We list the HR estimation accuracy of PRN trained with the proposed joint optimization pipeline (referred as PRN augmented), real samples (referred as PRN w/ Real), and synthetic samples (referred as PRN w/ Synth). The synthetic samples are generated by our generator $G$ through translating the real samples in the UBFC-RPPG training set when the joint optimization converges. As a comparison, we also include the performance of a state-of-the-art deep learning model 3D-CNN~\cite{tsou2020multi} that is trained with both real and synthetic samples (referred as 3D-CNN w/ Real$\&$Synth), just real samples (referred as 3D-CNN w/ Real), and just synthetic samples (referred as 3D-CNN w/ Synth). Performance of three traditional methods (POS~\cite{wang2016algorithmic}, CHROM~\cite{de2013robust} and ICA~\cite{poh2010advancements}) are also provided in the table.

Notably, the proposed PRN architecture has already outperformed other rPPG estimation methods even without synthetic skin color augmentation. More specifically, the proposed PRN has around $31\%$ improvement on MAE and around $14\%$ improvement on RMSE over the state-of-the-art 3D-CNN using real training samples. With the synthetic augmentation, the performance of PRN can be further improved. PRN trained with augmentation achieves $9\%$ improvement on MAE (from 0.75 BPM to 0.68 BPM) as compared with PRN trained with just real samples. This suggests that even for UBFC-RPPG dataset which is overwhelmed by subjects with light skin tones, increasing the diversity of training samples is still able to enhance the performance. This finding is consistent with the recent research~\cite{larrazabal2020gender} that demonstrates a balanced dataset can lead to optimal performance for all the groups. 

The joint optimized generator $G$ can be beneficial to other data-driven models as well. We train 3D-CNN with both real and corresponding synthetic samples from $G$. As compared with the 3D-CNN model trained with just real samples, 3D-CNN model trained with both real and synthetic samples exhibits $18\%$ improvement on MAE and $13\%$ improvement on RMSE. This further indicates that our generator has successfully learned to produce both visually-satisfying and BVP-informative facial videos, and these synthetic videos can facilitate the learning progress of the existing data-driven rPPG estimation algorithm without conducting the joint optimization process again to adapt to another new network architecture.

\subsection*{Cross-dataset performance on VITAL} \label{sec:vital} 

\begin{figure*}[t]
 \footnotesize
\begin{minipage}[b]{0.33\linewidth}
  \begin{tabular}{lccccccccc}
    \toprule
    \multirow{2}{*}{Method} & \multicolumn{2}{c}{\centering F1-2} &\multicolumn{2}{c}{\centering F3-4}&\multicolumn{2}{c}{\centering F5-6} &\multicolumn{2}{c}{\centering Overall}\\
    \cmidrule(l){2-3} \cmidrule(l){4-5} \cmidrule(l){6-7} \cmidrule(l){8-9} 
    ~ & MAE & RMSE & MAE & RMSE  & MAE & RMSE & MAE & RMSE\\
    \midrule
  PRN augmented & 2.37 & 3.13 & \textbf{2.95} & \textbf{3.82} & \textbf{4.97} &\textbf{6.83} & \textbf{3.16} & \textbf{4.19}\\
    
  PRN w/ Real  & 3.38 & 4.46 &4.67 & 5.83 & 6.92&8.96 & 4.67 & 5.98\\
PRN w/ Synth & 4.36 & 6.19 & 4.52 & 6.18 &5.61 & 8.15 & 4.69 & 6.59\\
    3D-CNN~\cite{tsou2020multi} w/ Real\&Synth & \textbf{2.32} & \textbf{3.11} &3.18 & 4.09 & 5.89&7.83&3.43 &4.51\\
  3D-CNN~\cite{tsou2020multi} w/ Real & 3.31 &4.64 &5.86 & 6.78 &7.19&9.02 &5.21 & 6.47\\
    3D-CNN~\cite{tsou2020multi} w/ Synth & 3.88 & 5.23 &4.00 & 5.71 & 6.34 &8.35& 4.44 &6.08\\
  \hdashline
  POS~\cite{wang2016algorithmic} & 6.20 &7.56 &7.80 & 9.20 &6.90 &9.01 & 7.03 & 8.57\\
  CHROM~\cite{de2013robust} &6.02 & 7.39 &7.01 & 8.33 & 6.93 &8.22 & 6.64 & 7.97\\
  ICA~\cite{poh2010advancements} &  7.72 &8.64 &9.57 & 10.82 & 6.74 & 8.07 & 8.31 & 9.46\\
  \bottomrule
\end{tabular} \label{tab:vital}
\end{minipage}\hfill
\begin{minipage}[b]{0.33\linewidth}
    \raisebox{-\height}{\includegraphics[width=\columnwidth]{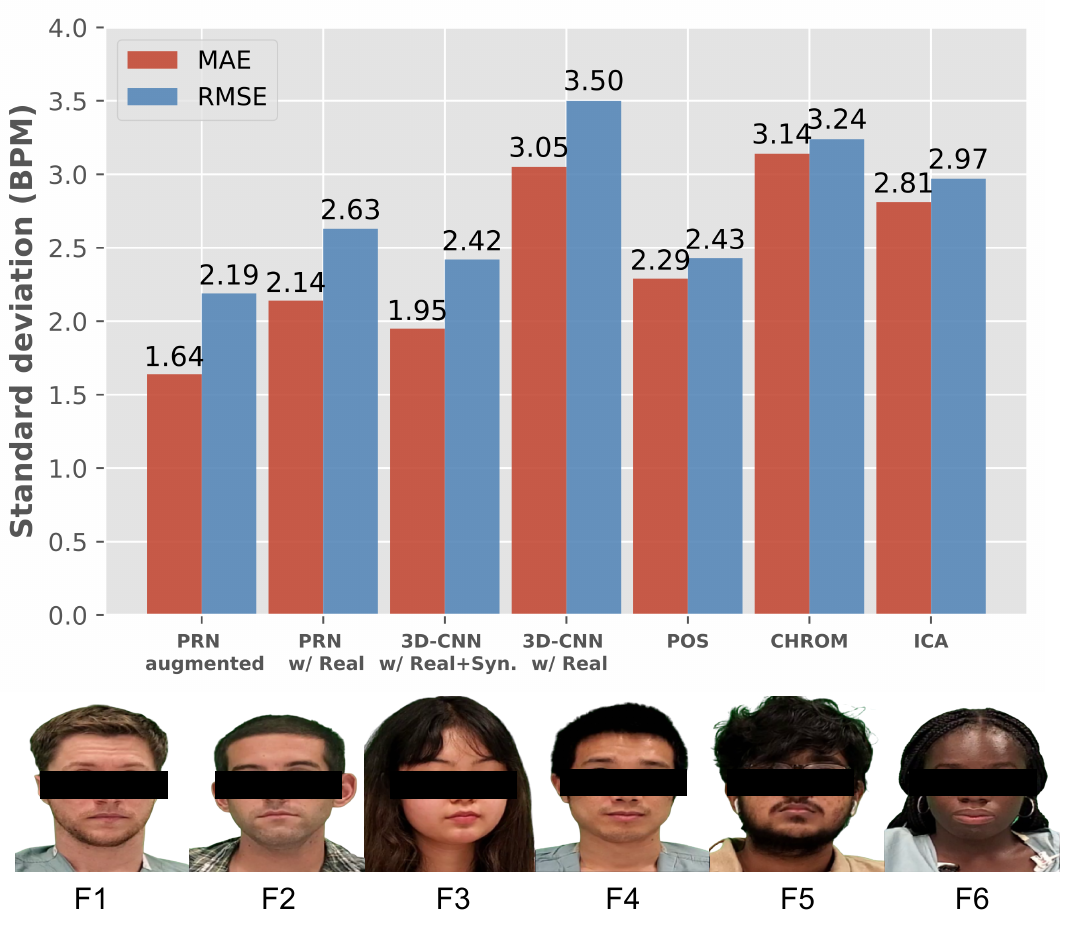}}
    \label{fig:std}
\end{minipage}

\caption{\textbf{Left: The proposed method shows an improved HR estimation accuracy on the VITAL dataset.} Boldface font denotes the preferred results. \textbf{Right: Synthetic dark-skinned videos can help to reduce bias in HR estimation.} The augmented PRN and the 3D-CNN~\cite{tsou2020multi} trained on both real and synthetic videos show a reduced standard deviation on MAE and RMSE across Fitzpatrick scales F1-6 in the VITAL dataset.} \label{fig:vital}
\end{figure*}

In real-world applications, it is common that the test subjects are in a different environment (e.g., illumination conditions) in contrast to the training samples. Therefore, we conduct a cross-dataset evaluation on the VITAL dataset using the models trained on the UBFC-RPPG videos. This type of cross-dataset verification can provide more visibility on the generalization capability of the models. Similarly, we report MAE and RMSE of various models trained with real and synthetics samples as shown in Figure~\ref{fig:vital}. Since VITAL dataset contains testing subjects of diverse skin tones 
with the associated Fitzpatrick scale labels (F1-6), we group the subjects into three categories, i.e., F1-2 (light skin color), F3-4 (medium skin color), and F5-6 (dark skin color), to measure the performance across different demographic groups. Please refer to the supplementary material for the SNR and PCC metrics.

PRN trained with the joint optimization pipeline exhibits significant improvement across these metrics as compared with PRN trained with just real samples. More precisely, there is 1.01 BPM reduction on MAE and 1.33 BPM reduction on RMSE for the light skin color group, 1.72 BPM reduction on MAE and 2.01 BPM reduction on RMSE for the medium skin color group, and 1.95 BPM reduction on MAE and 2.13 BPM reduction on RMSE for the dark skin color group. For all the methods, it is observed that the error of light skin tone group is generally lower than other groups. This is probably due to the melanin concentration of the light-skinned subjects is the least, and more light can be reflected to the camera. However, it should also be noted that models trained by both real and synthetic data have a relatively smaller performance difference among the three groups. For the medium and dark skin color groups, PRN trained with synthetic data shows lower estimation errors as compared with real data, and the errors are reversed for the light skin color group. This validates the fact that data-driven rPPG estimation models are heavily impacted by the skin color distribution of training samples, and it is critical to create a diverse and balanced training set for generalizability and real-world deployment of rPPG algorithms.

To assess the cross-dataset generalization capability of synthetic videos, we also evaluate 3D-CNN trained on real and synthetic samples from UBFC-RPPG on the VITAL dataset. Similar improvement can be observed in the 3D-CNN model, where 3D-CNN trained with both real and synthetic samples outperforms the model trained on only real or only synthetic samples. This supports that our synthetic videos can accurately reflect subtle color variations due to blood volume changes and can serve as a bio-realistic augmentation to the real samples. 

POS~\cite{wang2016algorithmic}, CHROM~\cite{de2013robust} and ICA~\cite{poh2010advancements} show relatively large HR estimation errors as compared with the data-driven models, where their MAEs on the light skin color group is usually larger than 6 BPM. Their MAEs are even higher for other groups. Unlike the end-to-end rPPG estimation networks, these conventional methods usually require preprocessing steps which may diminish the subtle color changes on the face and degrade the performance. Besides, these models need to average the pixel intensities over the skin region, and this might be a sub-optimal solution since skin pixels at different facial regions can contribute differently to the pulse signals. 

The cross-dataset experiment indicates that the improvement of our proposed framework is more substantial as compared with intra-dataset evaluation where all the samples are obtained within the same environment. This suggests that synthetic videos can provide more significant benefit by diversifying the training samples when there exist some data distribution shifts between real training and testing videos. This finding is also consistent with the observation for ray-tracing based augmentation method~\cite{mcduff2020advancing}. Synthetic augmentation techniques thus become particularly effective for cross-domain learning and can improve the generalization capability of HR estimation for real-world applications. 

\subsection*{Bias mitigation} \label{sec:bias}

It is critical for an algorithm to have consistent performance across different demographic groups in real-world medical deployment. To quantify the performance gap for each group, we use the standard deviation of MAE and RMSE for each Fitzpatrick scale as the measurement. This measurement has also been used in some prior work~\cite{mcduff2020advancing, yucer2020exploring}. The standard deviation for each method in the VITAL dataset is illustrated in Figure~\ref{fig:vital}, together with a sample portrait for each skin scale from F1 to F6. CHROM exhibits the largest variation (MAE: 3.14 BPM, RMSE: 3.24 BPM) across different Fitzpatrick scales, while the jointly optimized PRN shows the lowest bias (MAE: 1.64 BPM, RMSE: 2.19 BPM) as compared with all the conventional methods. In contrast to PRN trained with just real samples (MAE: 2.14 BPM), the augmented training offers a 23$\%$ improvement of bias mitigation among different groups while simultaneously improving the overall performance of all the groups. This suggests our joint training framework can provide a more desired trade-off between performance and bias. For 3D-CNN, the standard deviations for MAE and RMSE are also reduced by adding the synthetic samples into the training set. We attribute this improvement to the more diverse and balanced dataset augmented by our generator. 

\section*{Discussion}

In summary, it is worth noting that the lack of dark-skinned subjects in the existing rPPG datasets (MMSE-HR, AFRL, and UBFC-RPPG have roughly 10\%, 0\%, and 5\% dark-skinned subjects) has produced unwanted bias against some underrepresented groups, and there exist several practical constraints towards collecting a large-scale balanced dataset for rPPG. To address this issue, we propose a first attempt to translate facial frames from light-skinned subjects to dark skin tones while preserving the subtle color variations corresponding to the pulsatile signals. The proposed jointly optimized rPPG estimator can outperform the existing state-of-the-art methods with reduced estimation bias across different demographic groups. More specifically, PRN trained with augmentation has around 31\% reduction in MAE for the dark-skinned group along with 46\% improvement on bias mitigation in the VITAL dataset, as compared with 3D-CNN~\cite{tsou2020multi} trained with just real samples. Our generated synthetic videos maintain both photo-realistic and bio-realistic features and can be directly used to improve the performance of the existing deep learning rPPG estimation model. 

Our current pipeline is only a first attempt that focuses on the skin color translation, and all the remaining factors (e.g., pulse signals, body motion, and other facial attributes) are directly copied from the original videos. To maximize the benefit of synthetic augmentation, it is also critical to extend the generation framework to incorporate arbitrary facial attributes and pulse waves. We hope the method presented in this paper could inspire following work on synthetic generation for a more diverse dataset. Besides, it should also be noted that the generated frames are limited by a fixed resolution at $80 \times 80$. Future work may produce solutions to generate frames at arbitrary pixel resolution to fit the requirements of various subsequent rPPG estimation models without frame size interpolation. 

Video synthesis, such as deepfakes, has raised public concerns in the community~\cite{mirsky2021creation}. Over half a decade, these `fake' videos generated by deep learning have been used for face manipulation, and the malicious usage has drawn a lot of social attention. We demonstrate a positive example that these bio-realistic `fake' videos can also be utilized for the purpose of social good. Our synthetic videos are capable of reducing both HR estimation error and bias for rPPG models and further facilitate the development of remote healthcare. We hope our framework can act as a tool to address some social issues in the existing medical applications.

\section*{Methods}
\subsection*{Optical model for pulsatile blood variations}
In this section, we briefly review the existing skin reflection theory that models pulsatile blood variations. Under the assumption of a light source with a constant spectral composition and varying intensity, RGB channels $\mathbf{C}_k(t)$ at the $k$th skin pixel measured by a remote color camera can be described by the dichromatic reflection model as a time-varying function~\cite{wang2016algorithmic}:
\begin{equation} \label{eq:dichromatic_model}
    \mathbf{C}_k(t) = I(t) \cdot \big(\mathbf{v}_s(t) + \mathbf{v}_d(t)\big) + \mathbf{v}_n(t), 
\end{equation}
where $I(t)$ is the luminance intensity level, $\mathbf{v}_s(t)$ and $\mathbf{v}_d(t)$ are the time-varying specular and diffuse reflections respectively, and $\mathbf{v}_n(t)$ is quantization noise. Specular component $\mathbf{v}_s(t)$ in Equation~\eqref{eq:dichromatic_model} is a result of the mirror-like reflection from the skin surface, which is usually considered to be BVP independent. We can write $\mathbf{v}_s(t)$ as the following equation~\cite{wang2016algorithmic}:
\begin{equation} \label{eq:specular_component} 
    \mathbf{v}_s(t)=\mathbf{u}_s\cdot \big(s_0+s(t)\big),
\end{equation}
where $\mathbf{u}_s$ is the unit color vector of incident light, $s_0$ is the stationary part of the specular reflection, and $s(t)$ is varying part of the specular reflection induced by motion. Diffuse reflection $\mathbf{v}_d(t)$ in Equation~\eqref{eq:dichromatic_model} is related to the absorption and scattering properties of the skin tissues, and its varying component is identified as a key indicator to the blood volume changes~\cite{wang2016algorithmic}:
\begin{equation} \label{eq:diffuse_component}
    \mathbf{v}_d(t) = \mathbf{u}_d \cdot d_0 + \mathbf{u}_p \cdot p(t),
\end{equation}
where $\mathbf{u}_d$ is the unit color vector of the skin, $d_0$ is the stationary reflection strength, $\mathbf{u}_p$ is the relative pulsatile strengths in RGB channels, and $p(t)$ is the pulse signal. Substituting Equation~\eqref{eq:specular_component} and Equation~\eqref{eq:diffuse_component} into Equation~\eqref{eq:dichromatic_model}, we can write $\mathbf{C}_k(t)$ as follows:
\begin{equation} \label{eq:specual_and_diffuse}
     \mathbf{C}_k(t) = I(t) \cdot \Big(\mathbf{u}_s \cdot\big(s_0 +s(t)\big) + \mathbf{u}_d\cdot d_0 + \mathbf{u}_p \cdot p(t)\Big) + \mathbf{v}_n(t).
\end{equation}
The stationary parts of the specular and diffuse components can be combined into a single skin stationary term:
\begin{equation} \label{eq:combine_stationary}
    \mathbf{u}_c \cdot c_0 = \mathbf{u}_s \cdot s_0 + \mathbf{u}_d \cdot d_0,
\end{equation}
where $\mathbf{u}_c$ is the unit color vector of the skin reflection, and $c_0$ denotes the reflection strength. This further simplifies Equation~\eqref{eq:specual_and_diffuse} as:
\begin{equation} \label{eq:dichromatic_model_final}
    \mathbf{C}_k(t) = I_0 \cdot \big(1+i(t)\big) \cdot \big(\mathbf{u}_c \cdot c_0 + \mathbf{u}_s \cdot s(t) + \mathbf{u}_p \cdot p(t)\big) + \mathbf{v}_n(t), 
\end{equation}
where $I(t)$ is expressed as the sum of a stationary part $I_0$ and a time-varying motion-induced part $I_0 \cdot i(t)$. Video-based PPG measurement algorithms aim to estimate the pulse signal $p(t)$ from the pixel intensity $\mathbf{C}_k(t)$ by separating the physiological and non-physiological variations, while the primary focus of this paper is to establish an inverse mapping between $p(t)$ and $\mathbf{C}_k(t)$ for dark-skin realistic human faces in a data-driven manner.

\subsection*{Implementation details} \label{sec:implementation_details}
The facial bounding box for each video is estimated by applying a face detector based on Multitask Cascaded Convolutional Neural Networks (MTCNN)~\cite{zhang2016joint} to its first frame, and a square region with 160$\%$ width and height of the detected bounding box is cropped and resized to $80 \times 80$ using linear interpolation. The learning rate for the generator and the rPPG network are 0.0001 and 0.0003 respectively. The learning rates are modified base on a cosine annealing schedule during training~\cite{loshchilov2017sgdr}. The networks are initialized with Kaiming initialization~\cite{he2015delving} with a batch size of two and ReLU activation. We use Adam~\cite{kingma2014adam} solver with $\beta_1 = 0.5$ and $\beta_2 = 0.999$. The network architectures are implemented with batch normalization~\cite{ioffe2015batch} in Pytorch~\cite{NEURIPS2019_9015}, and the experiments are conducted on a single NVIDIA Tesla V100 GPU. 

\subsection*{Datasets} \label{sec:data}

\paragraph{UBFC-RPPG~\cite{bobbia2019unsupervised}:} UBFC-RPPG database contains 42 front facial videos from 42 subjects, and the corresponding ground-truth PPG singals are collected from a fingertip pulse oximeter. The videos are recorded at 30 frames per second with a resolution of 640x480 in the uncompressed 8-bit AVI format. Each video is roughly one minute long. 

\paragraph{VITAL dataset~\cite{chari2020diverse}:} 
Facial videos are recorded at 1920x1080 pixel resolution and 30 frames per second for 60 subjects at room lighting in the highly compressed MP4 format. Each video is roughly 2 minutes long. A Philips IntelliVue MX800 patient monitor is utilized for ground-truth vital sign monitoring. The subject wears a blood pressure cuff, 5-ECG leads, and a finger pulse oximeter, which is connected to the MX800 unit. Diverse skin tones and varied demographic groups are represented in the dataset. We use 58 subjects in the VITAL dataset (subject 26 and subject 40 are left out due to data errors in the collecting process). For the skin types quantified by Fitzpatrick scales~\cite{fitzpatrick1988validity}, there are 5, 16, 14, 11, 5, 7 subjects respectively from I (lightest) to VI (darkest). 

\paragraph{Comparison methods:} We compare our model with three conventional methods: POS~\cite{wang2016algorithmic}, CHROM~\cite{de2013robust} and ICA~\cite{poh2010advancements}. These rPPG baseline methods are implemented based on the publicly available MATLAB toolbox~\cite{mcduff2019iphys}, and we follow the procedures in the toolbox to obtain facial pixels of interest, i.e., converting facial frames from RGB to $YC_RC_B$ and identifying skin pixels based on a predefined threshold. We also compare with a data-driven state-of-the-art rPPG algorithm 3D-CNN~\cite{tsou2020multi}. It is implemented based on the architecture description as detailed in the original publication.

\paragraph{Evaluation metrics:} 

After obtaining the estimated pulse waves from each model, we apply a Butterworth filter to the output signals with cut-off frequencies of 0.7 and 2.5 Hz for heart rate estimation. The filtered waves are divided with sliding windows of 30-second length and 1-second stride, and a heart rate is estimated based on the position of the peak frequency for each window. For each subject, four error metrics are calculated and averaged over all windows. The four metrics include MAE, RMSE, PCC between the estimated hear rate and the ground-truth hear rate, and SNR of the estimated PPG waves. The ground-truth HR for UBFC-RPPG is obtained by applying the same procedures as described above to the ground-truth pulse waves, and the ground-truth HR for the VITAL dataset is obtained from the MX800 patient monitor through ECG signals. Please refer to the supplementary material for more details about these evaluation metrics. 

\section*{Acknowledgments}
AK acknowledges support from the National Science Foundation CAREER grant (IIS-2046737), the Army Young Investigator Program, and a Google Faculty Award.

\bibliographystyle{unsrt}  
\bibliography{references}  

\newpage

\begin{center}
    \LARGE\textbf{Supplementary Material}
    \vspace{0.5cm}
\end{center}

\appendix

\section{Error Metric Details}
To evaluate the heart rate (HR) estimation against the gold-standard ground truth, we use the following four metrics: mean absolute error (MAE), root mean square error (RMSE), Pearson’s correlation coefficient (PCC), and signal-to-noise ratio (SNR): 

\begin{align}
\centering
&\operatorname{MAE} = \frac{\sum_{i=1}^{N}\left|\mathrm{HR}_{i}-\mathrm{HR}_{i}\right|}{N}, \label{eq:mae}\\
&\mathrm{RMSE} = \sqrt{\frac{\sum_{i=1}^{N}\left(\mathrm{HR}_{i}-\mathrm{HR}_{i}\right)^{2}}{N}},\label{eq:rmse} \\
&\operatorname{PCC} = \frac{T \sum_{i=1}^{T} p_{i} \hat{p}_{i}  -\sum_{i=1}^{T} p_{i} \sum_{i=1}^{T} \hat{p}_{i}}{\sqrt{\left(T \sum_{i=1}^{T} p_{i}^{2}-\left(\sum_{i=1}^{T} p_{i}\right)^{2}\right)\left(T \sum_{i=1}^{T}\left(\hat{p}_{i}\right)^{2}-\left(\sum_{i=1}^{T} \hat{p}_{i}\right)^{2}\right)}}, \label{eq:pcc}\\
&\mathrm{SNR} = 10 \log _{10}\left(\frac{\sum_{f=0.75}^{2.5}\left(U_{t}(f) \hat{S}(f)\right)^{2}}{\sum_{f=0.75}^{2.5}\left(\left(1-U_{t}(f)\right) \hat{S}(f)\right)^{2}}\right), \label{eq:snr}
\end{align}
where $N$ is the total number of windows, $p$ is the ground-truth pulse wave, $\hat{p}$ is the estimated pulse signal, $\hat{S}$ is the power spectrum of the pulse signal, $f$ is the frequency in Hz, and $U_{t}(\cdot)$ is a binary mask. For the heart frequency region from $f_{\text{HR}}$ - 0.1 Hz to $f_{\text{HR}}$ + 0.1 Hz and its first harmonic region from 2 * $f_{\text{HR}}$- 0.1 Hz to 2 * $f_{\text{HR}}$ + 0.1 Hz, $U_{t}(\cdot)$ is set to be one. For other regions, $U_{t}(\cdot)$ is set be zero.

\section{Network Architecture}
We list the detailed architectures for the generator and the rPPG estimation network in this section.
\subsection{Generator}

The block diagram of the generator is illustrated in Figure~\ref{fig:resnetgenerator}. We adapt the architecture from the image translation networks in CycleGAN~\cite{zhu2017unpaired} and make the operation of 2D convolution to 3D convolution. The generator model consists of an encoder (several convolutional layers), a transformer (6 ResNet Blocks), and a decoder (several convolutional layers). 

\subsection{rPPG estimator} 
\begin{figure*}[t]
     \centering
     \begin{subfigure}[b]{.45\textwidth}
         \centering
         \includegraphics[width=\linewidth]{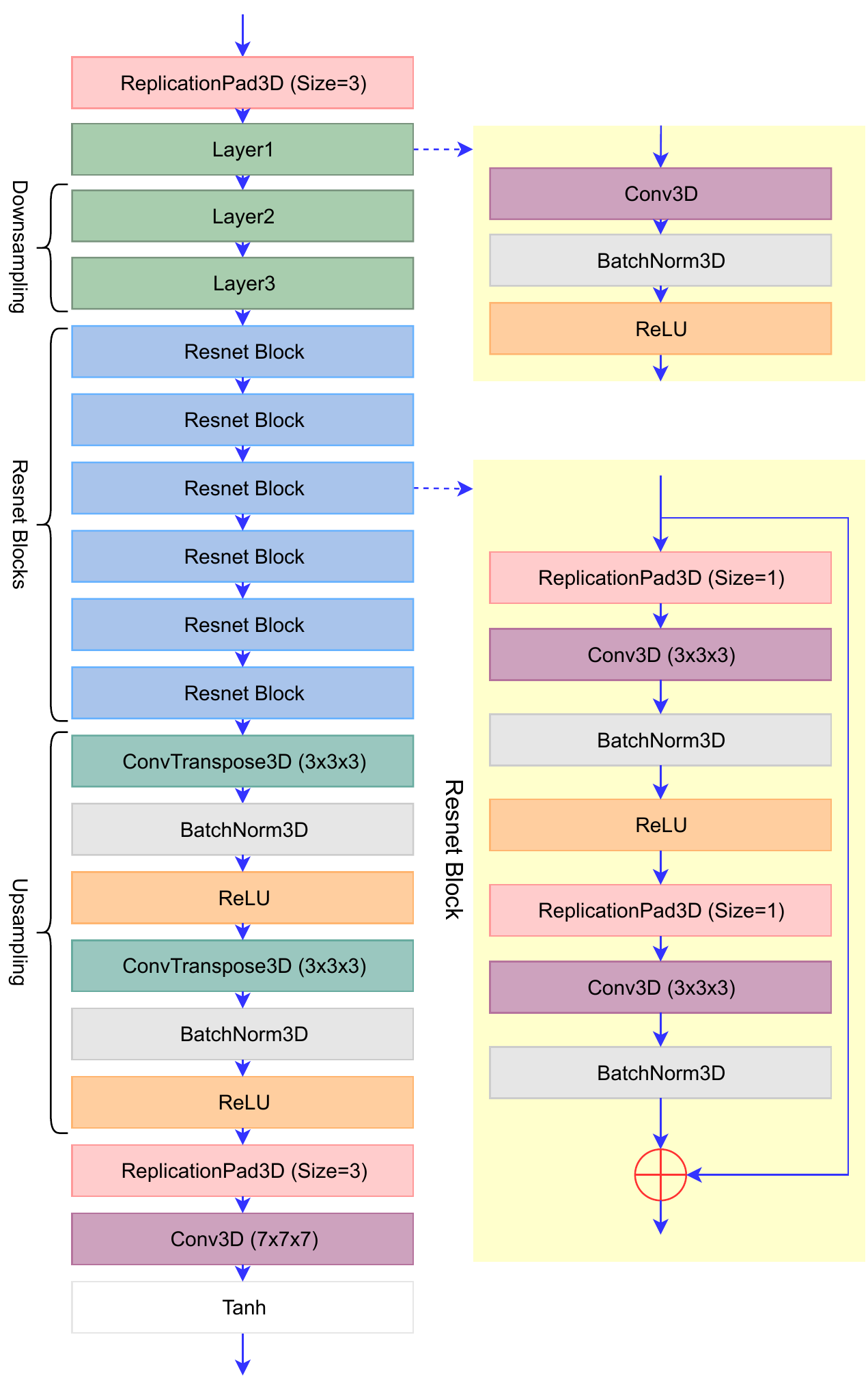}
         \caption{\textbf{Architecture of the generation network.}}
         \label{fig:resnetgenerator}
     \end{subfigure}
     \hfill
     \begin{subfigure}[b]{.45\textwidth}
         \centering
         \includegraphics[width=\linewidth]{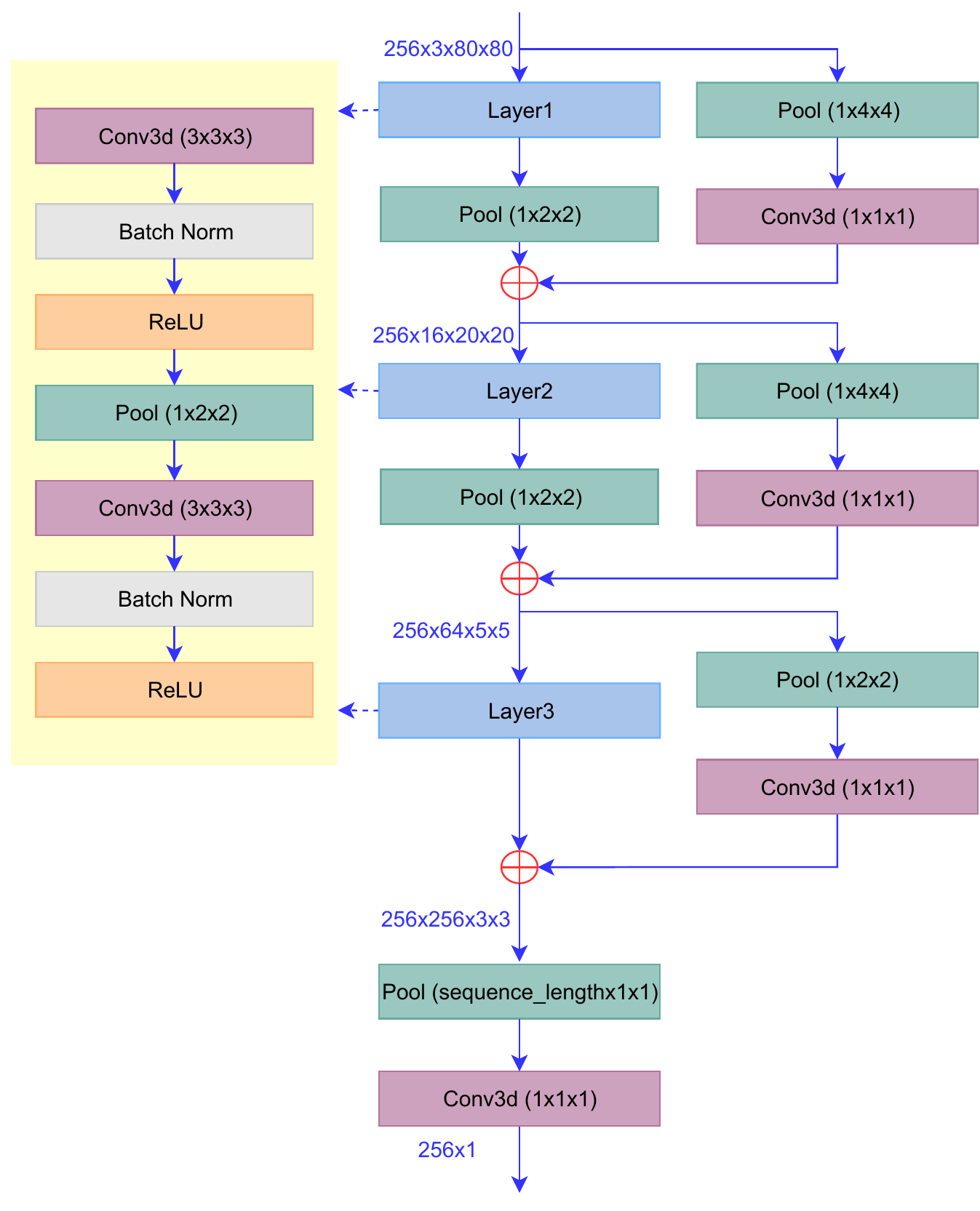}
         \caption{\textbf{Architecture of the rPPG estimation network.}}
         \label{fig:rppgnet}
     \end{subfigure}
     \caption{\textbf{Block diagrams of the networks used.}}
\end{figure*}

The diagram of the rPPG estimation network is shown in Figure~\ref{fig:rppgnet}. It consists of three consecutive 3D convolutional blocks with residual connections, and an average pooling is performed after each block for the downsampling purpose.

\section{More Results}
The additional results (PCC and SNR) of all the methods on VITAL dataset are provided in Table~\ref{tab:vital_more}. These experimental results on these two evaluation metrics are consistent with the results on MAE and RMSE and further validate the effectiveness of our proposed methods. 

\begin{table*}[h]
  \caption{\textbf{Performance of HR estimation on VITAL}. Boldface font denotes the best results.} 
    \label{tab:vital_more}
    \vskip 0.05in
  \begin{center}
  \begin{tabular}{lccccccccc}
    \toprule
    \multirow{2}{*}{Method} & \multicolumn{2}{c}{\centering F1-2} &\multicolumn{2}{c}{\centering F3-4}&\multicolumn{2}{c}{\centering F5-6}&\multicolumn{2}{c}{\centering Overall}\\
    \cmidrule(l){2-3} \cmidrule(l){4-5} \cmidrule(l){6-7} \cmidrule(l){8-9} 
    ~ & PCC & SNR & PCC & SNR & PCC & SNR & PCC & SNR\\
    \midrule
   PRN augmented & 0.40 & 3.45 & 0.63 & \textbf{5.73} & 0.22 & -4.24 & 0.46 & \textbf{2.84}\\
   PRN (w/ Real)  &0.36 & 0.32 & 0.50 & 0.03 & 0.04 & -6.79 & 0.35 & -1.28 \\
  PRN (w/ Synth) &0.29 &-0.45& 0.42 & -0.44 & 0.11 &-6.34 & 0.31 &-1.66\\
 3D-CNN~\cite{tsou2020multi} (w/ Real$\&$Synth) & \textbf{0.42} & \textbf{3.96} & \textbf{0.65} & 5.21 & \textbf{0.25} &-4.77&\textbf{0.48} &2.69\\
   3D-CNN~\cite{tsou2020multi} (w/ Real)  & 0.30 &-0.61 & 0.48 & -1.26&0.19&-8.10&0.35&-2.44\\
   3D-CNN~\cite{tsou2020multi} (w/ Synth) & 0.07 & -2.04 & 0.38 & -1.36&0.18&-5.82&0.23&-2.53\\
  \hdashline
  POS~\cite{wang2016algorithmic} & 0.16 & -1.31 & 0.36 & -0.78 & 0.09 & -4.50 & 0.23 & -1.74 \\
  CHROM~\cite{de2013robust} & 0.19 & -0.69 & 0.36 & -0.54 &-0.09 & \textbf{-4.22} & 0.21 & -1.36 \\
  ICA~\cite{poh2010advancements} &0.18 & -1.24 & 0.25 & -1.98 & 0.03 & -4.25 & 0.18 & -2.18 \\
    \bottomrule
\end{tabular}
\end{center}
\end{table*}

\end{document}